\pdfoutput=1

\documentclass[11pt]{article}

\usepackage[final]{acl}

\usepackage{times}
\usepackage{latexsym}
\usepackage{amsmath}
\usepackage{float}
\usepackage{tcolorbox}
\usepackage{array}
\usepackage[T1]{fontenc}

\usepackage[utf8]{inputenc}

\usepackage{microtype}

\usepackage{inconsolata}
\usepackage{graphicx}
\usepackage{multirow}

\usepackage{booktabs}
\usepackage{threeparttable}  
\usepackage{pifont} 

\newcommand{\cmark}{\textcolor{green!60!black}{\ding{51}}} 
\newcommand{\xmark}{\textcolor{red}{\ding{55}}}            

\title{\textsc{AutoForest}: Automatically Generating Forest Plots from Biomedical Studies with End-to-End Evidence Extraction and Synthesis}

\author{
 \textbf{Massimiliano Pronesti\textsuperscript{1,2}},
 \textbf{Angelo Miculescu\textsuperscript{2}},
 \textbf{Mohsin Kapdi\textsuperscript{2}},
 \textbf{Paul Flanagan\textsuperscript{2}},\\
 \textbf{Oisín Redmond\textsuperscript{2}},
  \textbf{Joao Bettencourt-Silva\textsuperscript{1}},
   \textbf{Gurdeep S. Mannu\textsuperscript{4}}, \textbf{Spiros Denaxas\textsuperscript{3,5}}, \\
   \textbf{Rui Bebiano Da Providencia E Costa\textsuperscript{3}},
 \textbf{Anya Belz\textsuperscript{2}},
 \textbf{Yufang Hou\textsuperscript{1,5}}
\\
 \textsuperscript{1}IBM Research
 \textsuperscript{2}Dublin City University
 \textsuperscript{3}UCL \textsuperscript{4}University of Oxford \\
 \textsuperscript{5}IT:U Interdisciplinary Transformation University Austria
\\
 \small{
   \textbf{Correspondence:} \href{mailto:massimiliano.pronesti@ibm.com}{massimiliano.pronesti@ibm.com}, \href{mailto:yufang.hou@it-u.at}{yufang.hou@it-u.at}
 }
}

\begin{document}
\maketitle

\begin{abstract}
Systematic reviews rely on forest plots to synthesise quantitative evidence across biomedical studies, but generating them remains a fragmented and labour-intensive process. Researchers must interpret complex clinical texts, manually extract outcome data from trials, define appropriate interventions and comparators, harmonise inconsistent study designs, and carry out meta-analytic computations—typically using specialised software that demands structured inputs and domain expertise. While recent work has demonstrated that large language models can extract study-level data from unstructured text, no existing system automates the complete pipeline from raw documents to synthesised forest plots. To address this gap, we introduce \textsc{AutoForest}\footnote{\sloppy\url{https://itu-nlp.github.io/projects/autoforest}}$^{,}$\footnote{\url{https://www.youtube.com/watch?v=R6ei97fOyXQ}}, the first end-to-end system that generates publication-ready forest plots directly from biomedical papers. Given one or more study papers, \textsc{AutoForest} automatically suggests ICO (Intervention, Comparator, Outcome) elements, extracts outcome data, performs statistical synthesis, and renders the final forest plot. 
We describe the system architecture, user interface and demonstrate its effectiveness on real-world examples through a user study involving clinicians, showing how \textsc{AutoForest} can accelerate evidence synthesis and substantially lower the barrier to conducting meta-analyses. 
\end{abstract}

\section{Introduction}

\begin{figure}[t!]
    \centering
    \includegraphics[width=0.9\linewidth]{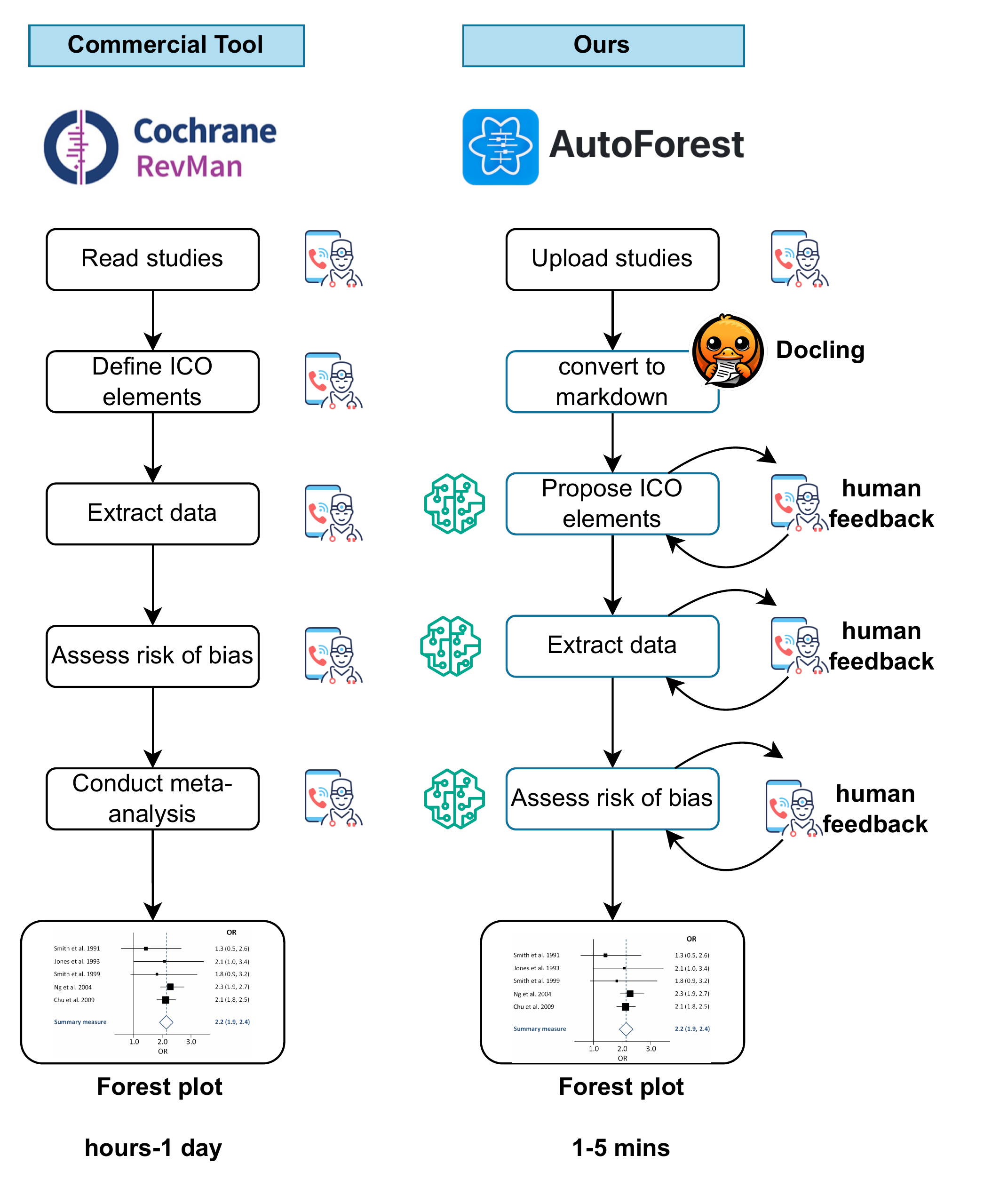}
    \caption{\textsc{AutoForest} automatically generates forest plots in minutes instead of hours with minimal human effort, as opposed to existing commercial tools.}
    \label{fig:enter-label}
\end{figure}

Systematic reviews are essential to evidence-based medicine, combining results from multiple biomedical studies to answer clinical questions with increased statistical power and reduced uncertainty. A key tool in this process is the forest plot, which visualises the estimated effects and confidence intervals across studies, enabling transparent comparison and meta-analysis. Despite its importance, generating forest plots remains a manual and time-consuming task. As seen in Figure~\ref{fig:enter-label}, researchers must first complete several upstream steps, including extracting outcome data from primary studies—often in unstructured PDF formats—defining the relevant intervention, comparator, and outcome (ICO) elements, and harmonising study designs and measurements. Following these steps, researchers use specialised software such as RevMan \cite{RevManManual}, which is used to conduct statistical synthesis using meta-analytic models and generate the final forest plot. RevMan requires users to manually enter data after converting it to a specific structured format (e.g.\ event counts and sample sizes for intervention and control groups); it offers no support for unstructured inputs or automation of the upstream steps.

 Recent research has made progress in automating individual components of this workflow. For example, large language models have been used to extract numerical results from clinical trial reports~\cite{yun2024automatically, sun2024good, lai2025language}, and to support reasoning about study-level effects~\cite{pronesti-etal-2025-enhancing} and risk of bias assessment~\cite{ji-etal-2025-robguard, wang-etal-2025-measuring, pronesti2026outcomeverificationverifiableprocess}. The approach by~\citet{pronesti-etal-2025-enhancing} introduces custom models paired with reinforcement learning to identify relevant numerical outcomes and estimate treatment effects across studies. However, it stops short of generating full meta-analytic visualisations, and still relies on manually specified ICO elements and downstream analysis tools. Other efforts employ interactive AI agents to streamline systematic review processes by assisting with study selection and summary generation, but they do not extract numerical effect sizes, perform statistical aggregation, or generate plots~\cite{qiu2025completingsystematicreview}.

To address these gaps, we introduce \textsc{AutoForest}, the first end-to-end system that produces publication-ready forest plots directly from biomedical papers. Given one or more PDF documents, \textsc{AutoForest} automatically suggests ICO elements, extracts outcome data using a numerical reasoning module, performs statistical synthesis using a random or fixed-effects model, and renders the resulting forest plot—all with minimal user input. Unlike traditional tools such as RevMan which require manual data extraction and entry, \textsc{AutoForest} uses a unified pipeline to automatically extract data directly from unstructured text and perform statistical synthesis and bias assessment.

We demonstrate the system on real-world reviews and show that it can recover forest plots with quality comparable to expert-curated outputs. By automating the full workflow from document ingestion to quantitative synthesis, \textsc{AutoForest} drastically reduces the time and expertise required to perform meta-analysis, making high-quality evidence synthesis more accessible and scalable.

\section{Background and Related Work}

\paragraph{Systematic Reviews} \hspace{-.2cm}are widely regarded as the gold standard in evidence-based medicine, providing rigorous syntheses of research to guide clinical decision-making \cite{murad2016}. They aim to address the challenge of staying up-to-date with an ever-growing volume of medical literature by aggregating high-quality evidence for specific clinical questions \cite{higgins2024}. However, producing a systematic review is 
time-consuming and costly: a 2019 study estimated that on average it takes 1–2 years and over \$141,000 to complete one \cite{michelson2019}. Given the substantial resources required, there is growing interest in automating various steps of the process \cite{marshall2019,khraisha2024,yun2024automatically, wang2024accelerating, pronesti-etal-2025-enhancing}.

\paragraph{Forest Plots} \hspace{-.2cm}constitute the cornerstone of quantitative synthesis in systematic reviews, visually summarising effect sizes and their associated confidence intervals across studies \cite{higgins2024}. Each study in a forest plot is represented by a point estimate and confidence interval, facilitating immediate visual assessment of treatment effects and heterogeneity across studies. Despite their widespread use, generating forest plots is often challenging due to inconsistencies in study reporting, the need for manual data extraction from narrative texts and tabular data, and the statistical computations required to produce aggregated results \cite{pronesti-etal-2025-query, yun2024automatically, pronesti-etal-2025-enhancing}.


\paragraph{Related Works.}
\begin{table*}[t]
\centering
\footnotesize
\begin{tabular}{lccccc}
\toprule
\textbf{Feature} & \textbf{RevMan} & \textbf{CMA} & \textbf{DistillerSR} & \textbf{EPPI-Reviewer} & \textbf{AutoForest (Ours)} \\
\midrule
Process raw documents                         & \xmark & \xmark & \cmark & \cmark & \cmark \\
ICO suggestion                          & \xmark & \xmark & \xmark & \xmark & \cmark \\
Automatic data extraction               & \xmark & \xmark & \xmark & \xmark & \cmark \\
Automatic risk of bias assessment & \xmark & \xmark & \xmark &  \xmark & \cmark \\
Forest plot generation                  & \cmark & \cmark & \xmark & \cmark & \cmark \\
Full pipeline automation                & \xmark & \xmark & \xmark & \xmark & \cmark \\
\bottomrule
\end{tabular}
\caption{Comparison of \textsc{AutoForest} with existing meta-analysis tools across key steps in the evidence synthesis workflow. \cmark{} indicates full support/automation, \xmark{} indicates no support.}
\label{tab:comparison}
\end{table*}

Current tools for meta-analytic synthesis, such as RevMan~\cite{RevManManual}, primarily provide interfaces for manual data entry, requiring structured inputs and extensive human intervention at all upstream stages, including ICO (Intervention, Comparator, Outcome) identification, data extraction and risk of bias estimation. Commercial tools like Comprehensive Meta‑Analysis (CMA)\footnote{\url{https://www.meta-analysis.com}} offer streamlined GUI workflows for entering structured data and generating forest plots, but rely entirely on manual ICO definition and data entry \cite{bax2007software,borenstein2022cma}. Web-based review platforms such as DistillerSR~\cite{distillersr2023ai} provide AI-assisted screening and extraction capabilities, but still require manual validation and external tools for plotting \cite{mcmillan2020drrs,distillersr2023ai}. EPPI‑Reviewer~\cite{distillersr2023ai} delivers integrated systematic review management with screening, extraction, and basic meta-analysis functions, yet again depends on user-driven ICO configuration and dataset curation \cite{thomas2010eppi}. In addition, risk of bias assessment (RoB)—an essential component of evidence synthesis—remains largely manual and delegated to separate tools. Established frameworks such as RoB1~\cite{higgins2011rob1} and its successor RoB2~\cite{sterne2019rob2} provide structured criteria for evaluating bias in randomised trials, but their application is still guided by human reviewers, with current software offering little more than form-based interfaces for entering judgments rather than automating the reasoning process. 
A summary of these comparisons is provided in Table~\ref{tab:comparison}.

\section{\textsc{AutoForest}}

\begin{figure*}
    \centering
    \includegraphics[width=\linewidth]{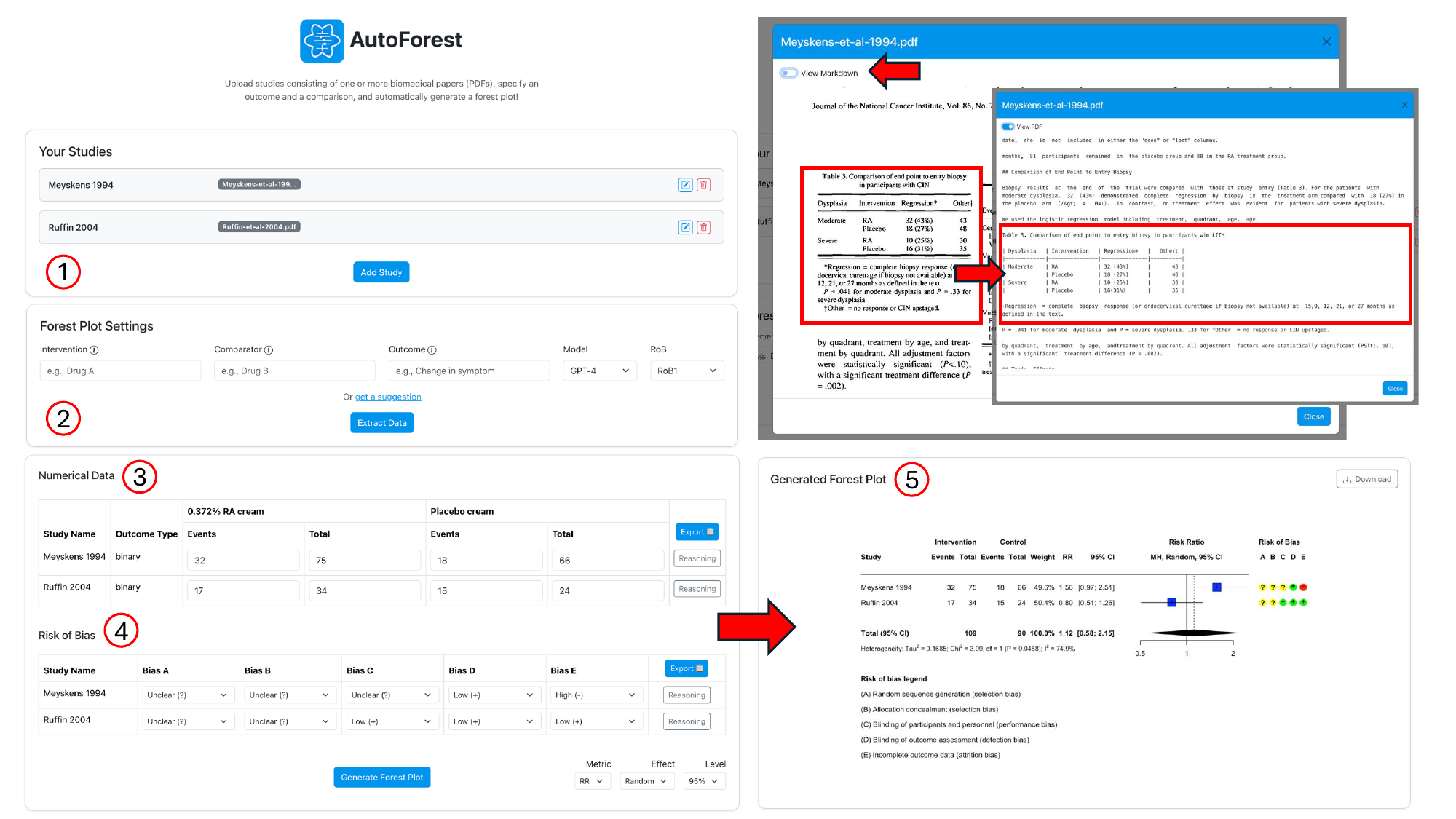}
    \caption{The interface of \textsc{AutoForest}. The user can upload their studies (one or more documents)~\textbf{(1)}, and generate ICO suggestions or manually input them~\textbf{(2)}. Data is extracted from every study based on the predicted outcome type and provided ICO triplet~\textbf{(3)}. Similarly, risk of bias is assessed based on the chosen RoB type (RoB1 or 2)~\textbf{(4)}. A forest plot is then generated based on the desired effect model and metrics~\textbf{(5)}.}
    \label{fig:interface}
\end{figure*}

\textsc{AutoForest} is an interactive system that enables users to extract, validate, and visualise results from primary studies in the form of publication-ready forest plots. It is designed to support researchers conducting systematic reviews by automating time-consuming tasks such as evidence identification, numerical data extraction, risk of bias estimation, and meta-analytic aggregation, while ensuring interpretability and transparency of each step.

The architecture consists of a ReactJS frontend connected to a FastAPI backend. These components communicate through a thin REST API layer, and the backend delegates specific NLP and numerical tasks to model endpoints. Importantly, the system is model-agnostic: while we currently use large language models (LLMs) for evidence extraction and reasoning, the interface supports drop-in replacement of backend components, enabling experimentation with different models or pipelines.

\subsection{Overall Workflow}

Figure~\ref{fig:interface} depicts \textsc{AutoForest}'s interface. The user workflow is structured into five stages:

\begin{enumerate}
    \item Upload studies in PDF format.
    \item Define the comparison of interest as Intervention -- Comparator -- Outcome (ICO) triplets.
    \item Extract numerical evidence from each study.
    \item Assess risk of bias from each study for either  RoB1~\cite{higgins2011rob1} or RoB2~\cite{sterne2019rob2} domains.
    \item Generate and download a publication-ready forest plot.
\end{enumerate}

\noindent Each of these stages is implemented as a distinct component in the UI and internally, allowing 
users to iteratively refine earlier decisions.

\subsection{Document Upload and Parsing}

Users begin by uploading primary studies in PDF format. These typically correspond to clinical trials or observational studies. Upon upload, each document is converted into a structured format using the \texttt{docling} library~\cite{docling, livathinos2025doclingefficientopensourcetoolkit, livathinos2025advancedlayoutanalysismodels}, through which we render a side-by-side view of the original PDF and its Markdown representation (Figure~\ref{fig:interface}, top right). This dual view allows users to cross-reference raw and structured content when inspecting extracted evidence.

\subsection{ICO Selection}

To define the scope of the meta-analysis, users must specify an Intervention ($I$), Comparator ($C$), and Outcome ($O$) triplet. Rather than relying on manual entry or simplistic frequency heuristics, \textsc{AutoForest} employs a transformer-based language model to extract candidate ICO elements from each study (Prompt in Appendix~\ref{appendix:prompts}). Let $\mathcal{S} = \{s_1, \dots, s_n\}$ denote the set of input studies. For each $s_i \in \mathcal{S}$, we define a candidate ICO set as
\[
\mathcal{T}_i = \texttt{LLMExtract}(s_i) \subseteq \mathcal{I} \times \mathcal{C} \times \mathcal{O}
\]
where $\mathcal{I}, \mathcal{C}, \mathcal{O}$ are the sets of possible interventions, comparators, and outcomes, respectively. Each $\mathcal{T}_i$ may contain multiple plausible triplets. To derive a consistent meta-analytic scope, we compute the set-theoretic intersection:
\begin{align*}
\mathcal{T}_{\text{shared}} = \cap_{i=1}^n \mathcal{T}_i
\end{align*}
yielding the set of ICO triplets common to all studies. The user interface includes a ``Get a suggestion'' button that leverages $\mathcal{T}_{\text{shared}}$ to propose a candidate ICO scope that is compatible across all included studies and therefore suitable for meta-analytic comparison. Users may override or refine the selection before initiating evidence extraction. 

\subsection{Numerical Evidence Extraction}

Once an ICO triplet has been selected, \textsc{AutoForest} automatically extracts the numerical evidence necessary to compute effect sizes. Following the methodology described by \citet{pronesti-etal-2025-enhancing}, we distinguish between binary and continuous outcomes and apply tailored extraction logic for each.

For binary outcomes (e.g., \emph{number of deaths}), the system retrieves the number of events and the total sample size for each study arm.\footnote{In the context of a forest plot, an arm refers to a group of participants in a clinical trial receiving a specific treatment.} For continuous outcomes (e.g., \emph{mean blood pressure}), it extracts the means, standard deviations, and total sample sizes for both the intervention and comparator arms. These quantities are required to compute standard measures such as risk ratios or mean differences used in meta-analysis.

Extraction is carried out using a language model prompted with both the ICO definition and the full text of each study (Appendix~\ref{appendix:prompts}). For every study, the model proposes the relevant numerical values along with a free-text explanation of how those numbers were identified. All extracted values are directly editable in the interface and exportable in YAML format. This design supports a human-in-the-loop workflow, where researchers can inspect, validate, and override model outputs. The accompanying reasoning text helps identify potential misinterpretations and provides transparency of the model’s decision-making process. Users may revise any extracted values, ensuring that the final extractions reflect expert judgement while benefiting from automation to reduce manual effort. The extraction step results in a structured evidence table where each row corresponds to a study and contains the extracted statistics and associated justifications. 

\subsection{Risk of Bias Assessment} Similarly to numerical data extraction, RoB assessment is performed by inputting the ICO triplet, the study text, and the risk domains under evaluation (prompt in Appendix~\ref{appendix:prompts}), offering support for both RoB1~\cite{higgins2011rob1} and RoB2~\cite{sterne2019rob2}. The model produces domain-level judgments along with a domain-level explanation pointing to the underlying study text. 

In addition, following the methodology described in~\citet{pronesti2026outcomeverificationverifiableprocess}, we support a step-level workflow for RoB2 in which the model is tasked with answering the signalling questions defined within each domain rather than directly assigning the final judgment. These question-level responses are subsequently processed using the official RoB2 decision algorithms (“macros”~\cite{sterne2019rob2}; Prompt in Figure~\ref{fig:rob2_macros_prompt}), allowing the system to derive the final risk-of-bias rating in a rule-based and fully transparent manner. This approach preserves model flexibility at the evidence-extraction stage while ensuring that the ultimate judgments remain consistent with RoB2 guidance.

\subsection{Forest Plot Generation}

After validating the evidence, users can generate a forest plot summarising the findings. \textsc{AutoForest} supports standard effect size measures (e.g., Risk Ratio, Mean Difference) and meta-analysis models (Fixed or Random Effects). We use the \texttt{meta} package in R~\cite{metaR} to ensure statistical correctness and compatibility with existing systematic review standards. The plot includes per-study point estimates and 95\% confidence intervals, pooled effect estimate, heterogeneity statistics (e.g., $I^2$ for  percentage of variation, $\tau^2$ for  variance of effects,  $p$-values). The resulting forest plot is displayed in the interface and can be exported in publication-quality formats. Examples generated by the system are shown in Appendix~\ref{appendix:comparison}.

\begin{table*}[h]
	\centering
	\footnotesize
	\begin{tabular}{llccccc}
		\toprule
		& & \multicolumn{2}{c}{\textbf{Data Extraction}} & \multicolumn{2}{c}{\textbf{RoB}} & \\
		\cmidrule(lr){3-4} \cmidrule(lr){5-6}
		\textbf{Group} & \textbf{Method} & \textbf{Acc (\%)} & \textbf{Edit rate (\%)} & \textbf{Acc (\%)} & \textbf{Edit rate (\%)} & \textbf{Time (min) $\downarrow$} \\
		\midrule
		Students
		& Manual (RevMan)               & 36.9  & -- &  40.0 & -- &  53.8\\
		& \textsc{AutoForest} only      &  83.3 & -- &  62.5 & -- &  --\\
		& \textsc{AutoForest} + edits   & \textbf{86.4}  &  2.8  &  \textbf{65.1} &  14.3  &  \textbf{26.5} \\
		\midrule
		Experts
		& Manual (RevMan)               & 45.8  & -- &  69.4 & -- &  70.4 \\
		& \textsc{AutoForest} only      &  82.5 & -- &   63.7& -- &  -- \\
		& \textsc{AutoForest} + edits   & \textbf{90.2}  & 10.4   & \textbf{79.2}  &  12.5  & \textbf{29.8}  \\
		\bottomrule
	\end{tabular}
    \begin{tablenotes}
    \footnotesize
    \item Wilcoxon signed-rank tests (full sample, $N=8$): Data Extraction  $p=0.013$, RoB  $p=0.050$, Time $p<0.001$.
    \end{tablenotes}
	\caption{Accuracy and change rate for data extraction and RoB tasks performed by students and domain experts using RevMan and \textsc{AutoForest}. ``\textsc{AutoForest} only'' indicates performance without human verification.}
	\label{tab:userstudy_results_multicol}
\end{table*}

\section{Evaluation}
\subsection{Experimental Setup}
We evaluate \textsc{AutoForest} through a combination of automatic evaluations and a controlled user study. Although the framework is model-agnostic and compatible with any LLM, our experiments use Claude Sonnet 4.5~\cite{sonnet4.5}, selected for its strong numerical reasoning and effectiveness in qualitative appraisal. Our goal is to assess the extent to which \textsc{AutoForest} accelerates and improves the production of forest plots, both when used fully automatically and in human-in-the-loop mode. The evaluation centers on 32 forest plots drawn from 18 Cochrane systematic reviews, covering a total of 56 included studies. For each plot, the task consists of extracting numerical evidence from raw full-text documents for a given ICO triplet and producing a publication-ready forest-plot entry. 
The study is guided by three research questions:  
\begin{itemize}
	\item \textbf{RQ1}: Does \textsc{AutoForest} reduce the time required to complete a forest plot compared to a manual workflow?
	\item \textbf{RQ2}: Does \textsc{AutoForest} improve the accuracy of forest plots relative to those created manually?
	\item \textbf{RQ3}: Can \textsc{AutoForest} help students achieve performance closer to that of domain experts?
\end{itemize}


\subsection{User Study}
We conducted a within-subjects user study involving four clinical domain experts and four graduate students familiar with the task
. Participants compared three workflows: a manual baseline (Manual + RevMan), \textsc{AutoForest} fully automated, and \textsc{AutoForest} AI-assisted (human-in-the-loop). Participants were instructed to extract the numerical and bias data required for each forest-plot entry, log their start and end times for each task, and—when using \textsc{AutoForest}—edit any incorrect numerical or RoB value.  After completing all tasks, participants rated the usability and efficacy of the tool on a 1-5 Likert scale. Full instructions provided to participants are available at this \href{https://autoforest.s3.eu-de.cloud-object-storage.appdomain.cloud/instructions_v4.pdf}{URL}.

\subsection{Results}
\paragraph{User Study} 

Results (Table~\ref{tab:userstudy_results_multicol}) demonstrate that \textsc{AutoForest} significantly outperforms manual workflows across all metrics. For \textbf{RQ1}, the tool nearly halved the time required to complete a forest plot for both groups ($p<0.001$). Regarding \textbf{RQ2}, the fully automated version (``\textsc{AutoForest} only'') achieved over 80\% accuracy in data extraction, a substantial improvement over the manual baselines. Human-in-the-loop edits further refined these results, reaching a peak accuracy for experts of 90.2\% for data extraction ($p = 0.013$) and 79.2\% for RoB ($p = 0.050$). For~\textbf{RQ3}, \textsc{AutoForest} significantly narrowed the performance gap between experience levels; students using the tool achieved 86.4\% accuracy, surpassing the manual performance of domain experts and nearly matching expert performance in the AI-assisted condition. 

Qualitative feedback (Table~\ref{tab:userstudy_usability}) indicates high system utility, with participants particularly valuing the transparency provided by the ``thought process". Overall, participants strongly endorsed the tool's potential for professional adoption (4.63/5), concluding that it could significantly reduce the human resources and time typically required to conduct systematic reviews (4.88/5).

\begin{table}[h]
	\centering
    \footnotesize
	\begin{tabular}
    {m{0.69\linewidth} >{\centering\arraybackslash}m{0.2\linewidth}}
		\toprule
		\textbf{Statement}                                    & \textbf{Rating} \\ 
		\midrule
		$\bullet$ The system was easy to use.                           &   4.63 $\pm$ 0.48   
        \\
		$\bullet$ The thought process in the data extraction step was useful to understand how each numerical value was derived and to verify or correct the automatically extracted data.                    & 4.25 $\pm$ 0.66  
        \\
		$\bullet$ The thought process in the risk of bias step was useful to understand the rationale behind each judgement and to verify or correct the automatically assigned risk of bias assessments.           &  4.13   $\pm$ 0.60 
        \\
		$\bullet$ The tool could be used in a professional context with minor improvements&   4.63 $\pm$ 0.48  
        \\
		$\bullet$ The tool could significantly reduce the time and number of professionals required to conduct a systematic review. & 4.88 $\pm$ 0.33
        \\
		\bottomrule
	\end{tabular}
	\caption{Post‐study usability ratings for the AI‐assisted conditions (1 = Strongly Disagree, 5 = Strongly Agree).}
	\label{tab:userstudy_usability}
\end{table}

\paragraph{ICO Suggestions} Let $I_{\mathrm{SR}} \in \mathcal{I}$, $C_{\mathrm{SR}} \in \mathcal{C}$, $O_{\mathrm{SR}} \in \mathcal{O}$ denote the sets of
interventions, comparators, and outcomes addressed in the systematic review,
and let $I_{\mathrm{sugg}}, C_{\mathrm{sugg}}, O_{\mathrm{sugg}}$ be the
corresponding sets suggested by the model. For each element type
$X \in \{I, C, O\}$, we compute: 
\begin{itemize}
	\item the proportion of correct suggestions (precision) 
	$P_X = \tfrac{|X_{\mathrm{sugg}} \cap X_{\mathrm{SR}}|}{|X_{\mathrm{sugg}}|}$
	\item the proportion of incorrect or hallucinated suggestions,
	$H_X = \tfrac{|X_{\mathrm{sugg}} \setminus X_{\mathrm{SR}}|}{|X_{\mathrm{sugg}}|}$
	\item  the proportion of elements in the systematic review that the model
	correctly captured (recall),
	$R_X = \tfrac{|X_{\mathrm{sugg}} \cap X_{\mathrm{SR}}|}{|X_{\mathrm{SR}}|}$
\end{itemize}
Results are reported in Table~\ref{tab:ico-eval}. We observe that the tool achieves high precision and strong coverage of the systematic review’s ICO elements, indicating accurate suggestions with limited omissions.

\begin{table}
	\centering
    \footnotesize
	\begin{tabular}{lccc}
		\toprule
		\textbf{Element} & $P_X\uparrow$ & $H_X\downarrow$ & $R_X\uparrow$  \\
		\midrule
		$I$ &  94.1 &  5.9 &  94.1 \\
		$C$ & 83.3 & 16.7 &  91.6  \\
		$O$ & 94.3 & 5.7 &   90.6\\
		\bottomrule
	\end{tabular}
	\caption{Evaluation metrics for suggested ICO elements.
		$P_X$: precision,
		$H_X$: hallucination rate,
		$R_X$: recall}
    	\label{tab:ico-eval}
\end{table}

\paragraph{Markdown Conversion} We assess the fidelity of the document-to-structure conversion layer used by \textsc{AutoForest} through 4 quantitative metrics capturing table detection, table parsing quality and numerical cell accuracy on the 206 tables contained in the 56 studies used for the user study (Table~\ref{tab:doc2struct}).

\begin{table}
	\centering
	\footnotesize
	\begin{tabular}{l c}
		\toprule
		\textbf{Metric} & \textbf{Acc (\%)} \\
		\midrule
		Table detection &  98.1  \\
		Table Structure (TEDS) & 93.4 \\
		Numerical Cell  & 97.8 \\
		Table captions & 98.1  \\
		\bottomrule
	\end{tabular}
	\caption{Document-to-structure conversion metrics.}
	\label{tab:doc2struct}
\end{table}

All metrics show that the conversion layer maintains table integrity and numerical fidelity with high accuracy, ensuring reliable downstream extraction.

\paragraph{RoB2 vs.\ RoB2+Macros}  
Table~\ref{tab:rob2_macro_comparison} reports the accuracy of the standard RoB2 workflow compared to the variant augmented with macro-based reasoning. The macros lead to a clear improvement, increasing overall domain-level accuracy from 61.6\% to 70.8\%. This confirms that incorporating structured reasoning templates helps the model produce more consistent and interpretable risk-of-bias judgments, in line with the findings of~\citet{pronesti2026outcomeverificationverifiableprocess}.

\begin{table}
	\small
	\centering
	\begin{tabular}{lc}
		\toprule
		\textbf{Method} & \textbf{Acc (\%)} \\
		\midrule
		RoB2 (standard) &  61.6 \\
		RoB2 (macros) &  \textbf{70.8} \\
		\bottomrule
	\end{tabular}
	\caption{Comparison of RoB2 evaluation with and without macro-based reasoning.}
	\label{tab:rob2_macro_comparison}
\end{table}



\section{Conclusions}
This paper introduced \textsc{AutoForest}, an end-to-end system designed to automate the generation of forest plots directly from unstructured biomedical study documents. The traditional process of creating these essential visualisations is labour-intensive, often taking several hours and requiring domain expertise. \textsc{AutoForest} addresses this bottleneck by integrating document parsing, automated ICO suggestions, evidence extraction, risk of bias assessment and reasoning to substantially reduce the time required to perform meta-analysis. Our user study 
suggests that \textsc{AutoForest} particularly when used in a human-in-the-loop capacity,
can accelerate evidence synthesis substantially in comparison to the standard manual workflow. Crucially, our results indicate that AI assistance may help bridge the gap between novice and expert performance; students using \textsc{AutoForest} outperformed the manual baseline of domain experts and approached the accuracy of experts using the tool. 

\section{Limitations}
While \textsc{AutoForest} demonstrates a significant reduction in the time and effort required to generate forest plots, we acknowledge several limitations in the current work.

First, the scope of our evaluation warrants a cautious interpretation of findings. While the user study provides valuable insights, we only conducted the experiment with a sample of 8 participants. A larger, more diverse cohort, with varied levels of expertise conducting systematic reviews, would be necessary to establish broader usability.  Moreover, the current evaluation focuses on standard parallel-group trial designs; complex designs such as multi-arm or crossover trials have not yet been explicitly tested in this paper. 


Second, \textsc{AutoForest} is designed to accelerate a specific segment of the evidence synthesis pipeline and does not address the full spectrum of tasks in a systematic review, such as literature searching, screening or study selection. 

Fianlly, while the system provides textual explanations to support its data extractions and risk of bias judgments, a current limitation is the absence of a visual feature to directly highlight this evidence in the source document. Implementing such a traceability feature would make checking the model's work significantly easier for the user and represents a valuable direction for future research. 


\bibliography{custom}

\clearpage 
\appendix

\section{Qualitative Comparison of Domain Experts with \textsc{AutoForest}}
\label{appendix:comparison}

\begin{figure}[H]
    \centering
    \includegraphics[width=\linewidth]{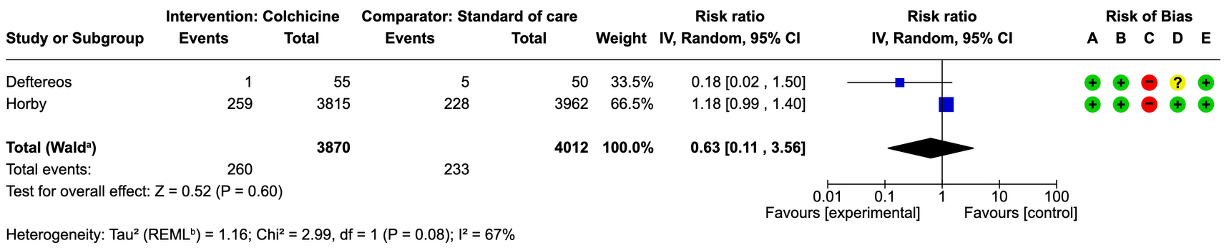}
    \caption{Forest plot created by a domain expert with RevMan. Only three of the eight numeric entries are correct (37.5\% Acc). The expert was unable to locate the relevant numerical data for one study and mistakenly used data from a different result section. The risk of bias table presents mistakes for bias B,C,D (50\% Acc).}
    \label{fig:de_fp}
\end{figure}

\begin{figure}[H]
    \centering
    \includegraphics[width=\linewidth]{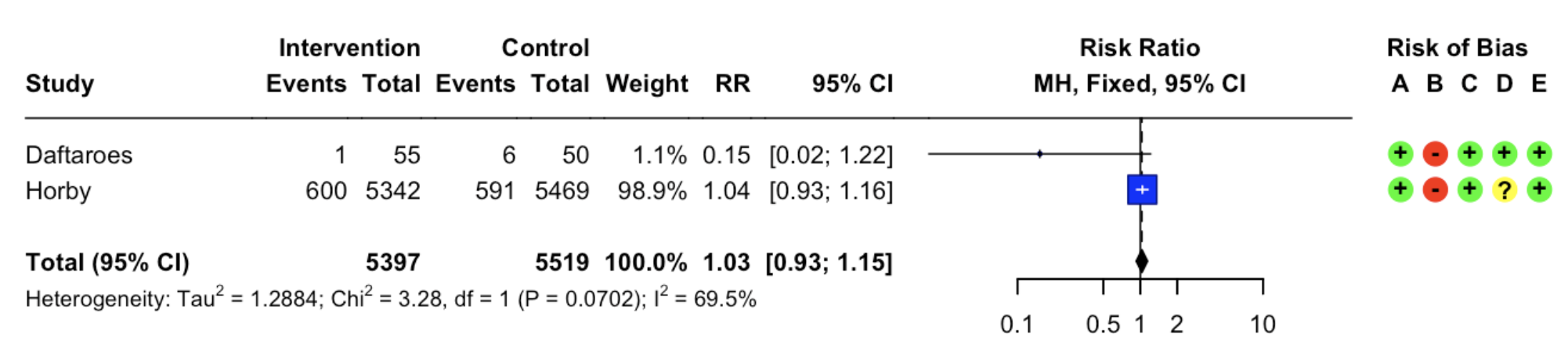}
   \caption{Forest plot generated by \textsc{AutoForest}. Out of eight numeric entries, only two are incorrect, demonstrating a substantial improvement in accuracy over the manual expert process (75\% Acc). The risk of bias map is all correct but bias D for the first study, also demonstrating a big leap with the manual version (90\% Acc).}
    \label{fig:autoforest_fp_1}
\end{figure}

\begin{figure}[H]
    \centering
    \includegraphics[width=\linewidth]{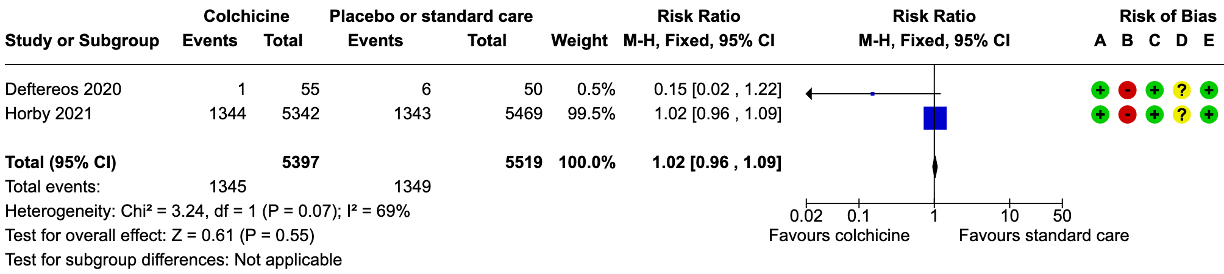}
    \caption{Ground truth forest plot.}
    \label{fig:gt_fp_1}
\end{figure}

We present a qualitative comparison of a forest plot manually created by a domain expert (Figure~\ref{fig:de_fp}) with the one generated by \textsc{AutoForest} with no human verification (Figure~\ref{fig:autoforest_fp_1}) and the ground truth from the original systematic review (Figure~\ref{fig:gt_fp_1}). 

\begin{figure}
    \centering
     \includegraphics[width=\linewidth]{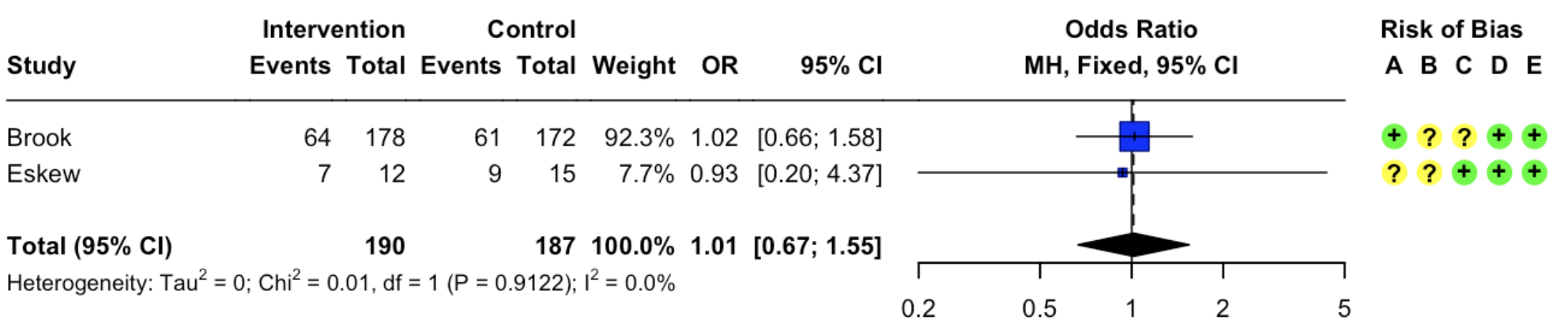}
    \includegraphics[width=\linewidth]{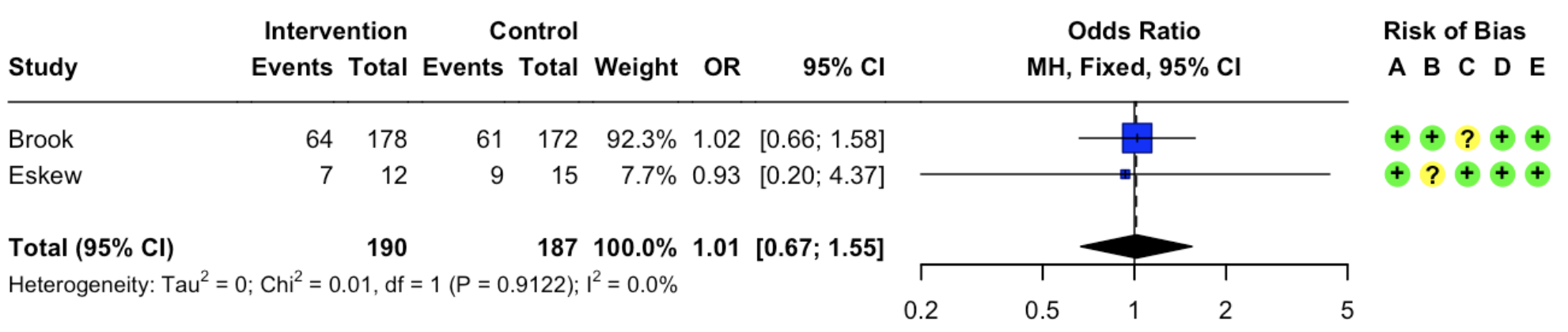}
    \includegraphics[width=\linewidth]{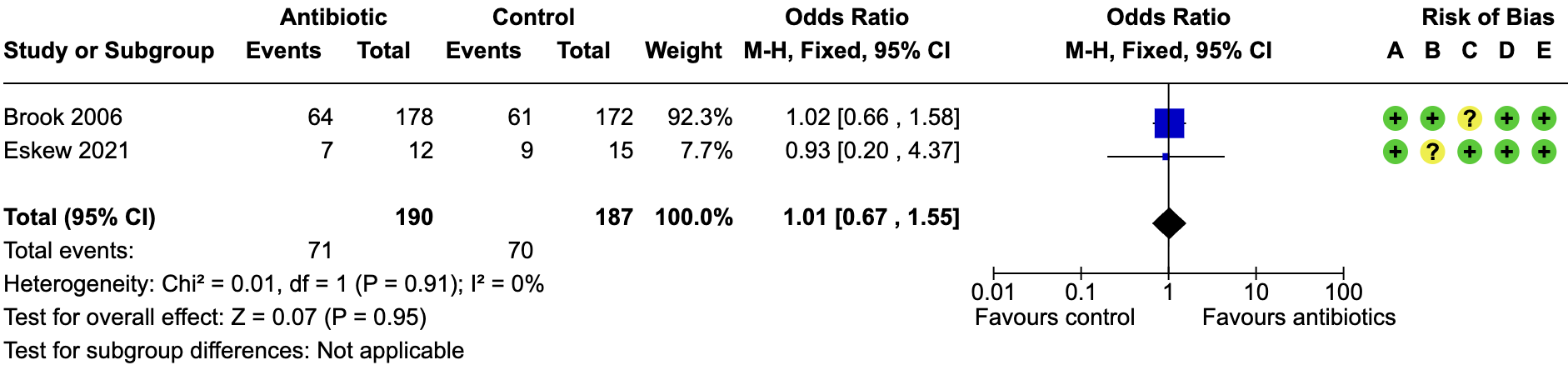}
    \caption{\textbf{Top}: Forest plot generated by \textsc{AutoForest}; \textbf{middle:} Forest plot edited by domain expert, scoring 100\% in both data extraction and risk of bias (RoB); \textbf{bottom:} ground truth from the original systematic review.}
    \label{fig:autoforest_fp_2_de}
\end{figure}

In addition, we show a comparison of forest plot generated by \textsc{AutoForest} (Figure~\ref{fig:autoforest_fp_2_de}, top) and further verified by a domain expert (Figure~\ref{fig:autoforest_fp_2_de}, middle) with the corresponding ground truth from the original systematic review (Figure~\ref{fig:autoforest_fp_2_de}, bottom).

These four visualisations illustrate the differences in accuracy across manual and automated forest plot construction. The domain expert, using RevMan, made several mistakes both in data extraction and risk of bias assessment. Most errors were due to difficulties locating the appropriate numerical values in the full text and selecting data from the wrong result section, or in the inner difficulty posed by qualitative appraisal in bias judgment. In contrast, \textsc{AutoForest} provided a very accurate starting point, suggesting that automated extraction can reliably support domain expert's workflow. When domain experts were provided with \textsc{AutoForest}'s output as a starting point, they were able to produce a fully accurate plot. This suggests that \textsc{AutoForest} can serve as a useful aid in manual review workflows.



\section{Prompts}\label{appendix:prompts}
\begin{figure}
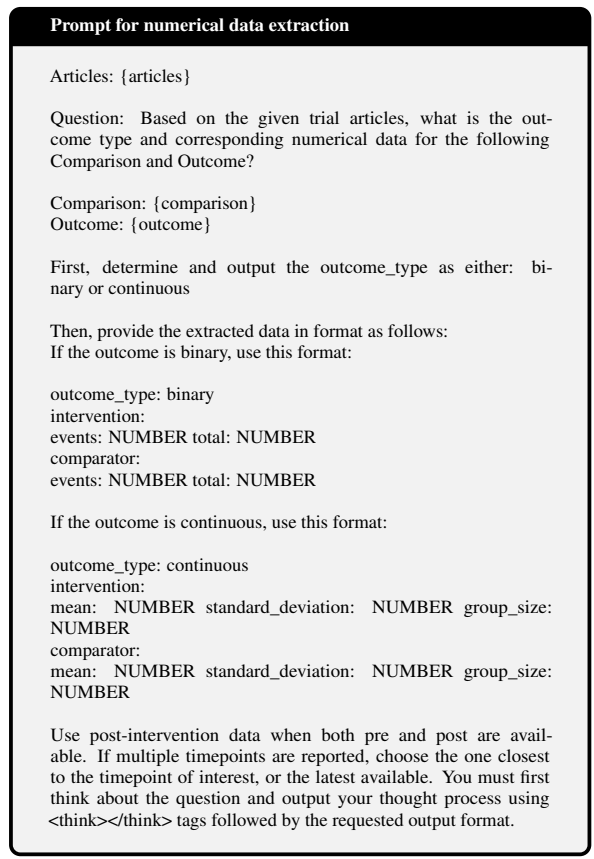

	\scriptsize
	\begin{tcolorbox}[
		colback=gray!10, 
		colframe=black, 
		boxrule=0.5mm,
		title={Prompt for numerical data extraction}, 
		fonttitle=\bfseries 
		]
		Articles: \{articles\}
		\\
            \\
		Question: Based on the given trial articles, what is the outcome type and corresponding numerical data for the following Comparison and Outcome?
		\\
		\\
		Comparison: \{comparison\}\\
		Outcome: \{outcome\}
		\\
		\\
		First, determine and output the outcome\_type as either: binary or continuous
		\\
		\\
		Then, provide the extracted data in format as follows:
		\\
		If the outcome is binary, use this format:
		\\
		\\
		outcome\_type: binary
		\\intervention:
		\\events: NUMBER total: NUMBER
		\\comparator:
		\\events: NUMBER total: NUMBER
		\\
		\\
		If the outcome is continuous, use this format:
		\\
		\\
		outcome\_type: continuous
		\\intervention:
		\\mean: NUMBER standard\_deviation: NUMBER group\_size: NUMBER
		\\
		comparator:
		\\mean: NUMBER standard\_deviation: NUMBER group\_size: NUMBER
		\\
		\\Use post-intervention data when both pre and post are available. If multiple timepoints are reported, choose the one closest to the timepoint of interest, or the latest available.
		You must first think about the question and output your thought process using <think></think> tags followed by the requested output format.
	\end{tcolorbox}
	\caption{Prompt for numerical data extraction.}
	\label{fig:data_extr_prompt}
\end{figure}

\begin{figure}
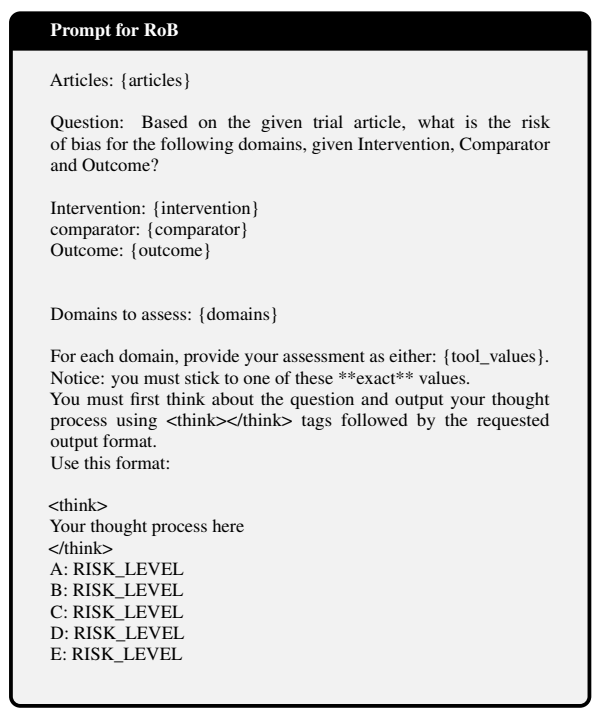

	\scriptsize
	\begin{tcolorbox}[
		colback=gray!10, 
		colframe=black, 
		boxrule=0.5mm,
		title={Prompt for RoB}, 
		fonttitle=\bfseries 
		]
		Articles: \{articles\}
        \\
		\\
        Question: Based on the given trial article, what is the risk of bias for the following domains, given Intervention, Comparator and Outcome?        \\
		\\
        Intervention: \{intervention\} \\
        comparator: \{comparator\} \\
        Outcome: \{outcome\} \\
        \\
        \\
        Domains to assess: \{domains\}
        \\
        \\
        For each domain, provide your assessment as either: \{tool\_values\}. \\
        Notice: you must stick to one of these **exact** values.

        You must first think about the question and output your thought process using <think></think> tags followed by the requested output format.\\
        Use this format:
        \\
        \\
        <think> \\
        Your thought process here\\
        </think>\\
        A: RISK\_LEVEL\\
        B: RISK\_LEVEL\\
        C: RISK\_LEVEL\\
        D: RISK\_LEVEL\\
        E: RISK\_LEVEL\\
	\end{tcolorbox}
	\caption{Prompt for Risk of Bias Assessment.}
	\label{fig:rob_prompt}
\end{figure}

\begin{figure}
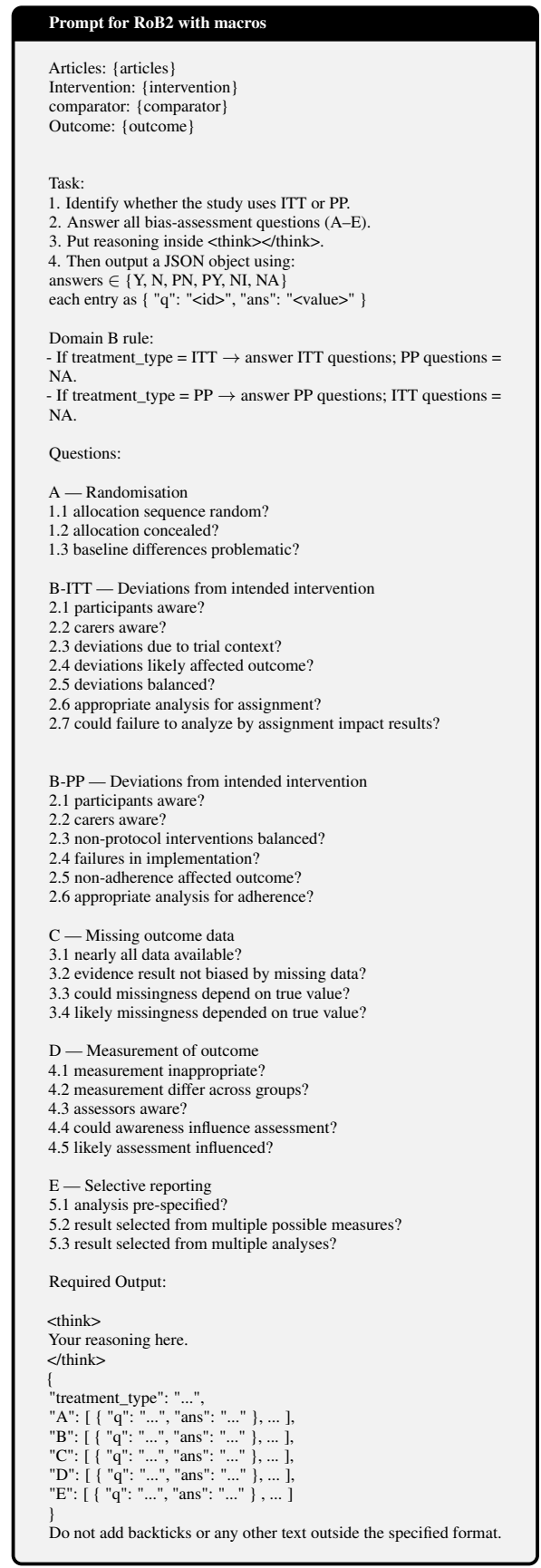

	\scriptsize
	\begin{tcolorbox}[
		colback=gray!10, 
		colframe=black, 
		boxrule=0.5mm,
		title={Prompt for RoB2 with macros}, 
		fonttitle=\bfseries 
		]
		Articles: \{articles\} \\
        Intervention: \{intervention\} \\
        comparator: \{comparator\} \\
        Outcome: \{outcome\} \\
        \\
        \\
        Task: \\
        1. Identify whether the study uses ITT or PP. \\
        2. Answer all bias-assessment questions (A–E). \\
        3. Put reasoning inside <think></think>. \\
        4. Then output a JSON object using: \\
            answers $\in$ \{Y, N, PN, PY, NI, NA\} \\
             each entry as \{ "q": "<id>", "ans": "<value>" \}
        \\
        \\
        Domain B rule:\\
          - If treatment\_type = ITT $\to$ answer ITT questions; PP questions = NA.\\
          - If treatment\_type = PP $\to$ answer PP questions; ITT questions = NA.
        \\
        \\
        Questions:
        \\
        \\
    A — Randomisation \\
      1.1 allocation sequence random? \\
      1.2 allocation concealed? \\
      1.3 baseline differences problematic? \\
    \\
    B-ITT — Deviations from intended intervention \\
    2.1 participants aware? \\
    2.2 carers aware? \\
    2.3 deviations due to trial context? \\
    2.4 deviations likely affected outcome? \\
    2.5 deviations balanced? \\
    2.6 appropriate analysis for assignment? \\
    2.7 could failure to analyze by assignment impact results? \\
    \\
    \\
    B-PP — Deviations from intended intervention \\
      2.1 participants aware? \\
      2.2 carers aware? \\
      2.3 non-protocol interventions balanced? \\
      2.4 failures in implementation? \\
      2.5 non-adherence affected outcome? \\ 
      2.6 appropriate analysis for adherence? \\
        \\
        C — Missing outcome data \\
  3.1 nearly all data available? \\
  3.2 evidence result not biased by missing data? \\
  3.3 could missingness depend on true value?\\
  3.4 likely missingness depended on true value?\\
\\
D — Measurement of outcome\\
  4.1 measurement inappropriate?\\
  4.2 measurement differ across groups?\\
  4.3 assessors aware?\\
  4.4 could awareness influence assessment?\\
  4.5 likely assessment influenced?\\
\\
E — Selective reporting\\
  5.1 analysis pre-specified?\\
  5.2 result selected from multiple possible measures?\\
  5.3 result selected from multiple analyses?\\
  \\
Required Output:
\\
\\
<think>\\
Your reasoning here.\\
</think>\\
\{\\
  "treatment\_type": "...",\\
  "A": [ \{ "q": "...", "ans": "..." \}, ... ], \\
  "B": [ \{ "q": "...", "ans": "..." \}, ... ], \\
  "C": [ \{ "q": "...", "ans": "..." \}, ... ],\\
  "D": [ \{ "q": "...", "ans": "..." \}, ... ],\\
  "E": [ \{ "q": "...", "ans": "..." \} , ... ]\\
\}
\\
Do not add backticks or any other text outside the specified format.
	\end{tcolorbox}
	\caption{Prompt for RoB2 with macros.}
	\label{fig:rob2_macros_prompt}
\end{figure}

The prompts used for the ICO suggestion, data extraction and risk of bias steps (with and without macros) are reported in Figure~\ref{fig:ico_prompt}, ~\ref{fig:data_extr_prompt}, ~\ref{fig:rob_prompt}, and~\ref{fig:rob2_macros_prompt} respectively. 

\begin{figure}[H]
	\scriptsize
	\begin{tcolorbox}[
		colback=gray!10, 
		colframe=black, 
		boxrule=0.5mm,
		title={Prompt for ICO suggestion}, 
		fonttitle=\bfseries 
		]
		Article: \{article\}
		\\
            \\
            From the text below, suggest the most likely values for intervention, comparator, and outcome as bullet lists under each heading.
Make sure every bullet is a self-explanatory word or short phrase that can be used as an ICO element in a meta-analysis.
\\
\\
Follow this format:\\
\\
\textasciigrave\textasciigrave\textasciigrave\\
Intervention:\\
- intervention 1\\
- intervention 2
...\\  
\\
Comparator:\\
- comparator 1\\
- comparator 2\\
- ...\\
\\
Outcome:\\
- outcome 1  \\
- outcome 2\\
...\\
\textasciigrave\textasciigrave\textasciigrave
\\
\\
Output:\\
	\end{tcolorbox}
	\caption{Prompt for ICO suggestion.}
	\label{fig:ico_prompt}
\end{figure}

\end{document}